%% file: main.tex
\documentclass[10pt,twocolumn,letterpaper]{article}

\usepackage{cvpr}              

\input{preamble}

\definecolor{cvprblue}{rgb}{0.21,0.49,0.74}
\usepackage[pagebackref,breaklinks,colorlinks,allcolors=cvprblue]{hyperref}

\def\paperID{} 
\def\confName{CVPR}
\def\confYear{2026}

\title{Neu-PiG: Neural Preconditioned Grids\\for Fast Dynamic Surface Reconstruction on Long Sequences}

\author{
Julian Kaltheuner \qquad Hannah Dröge \qquad Markus Plack \qquad
Patrick Stotko \qquad Reinhard Klein\\
University of Bonn\\
{\tt\small \{kaltheun,droege,mplack,stotko,rk\}@cs.uni-bonn.de}
}

\begin{document}


\twocolumn[{%
\renewcommand\twocolumn[1][]{#1}%
\maketitle
\centering
\includegraphics[width=0.92\linewidth, trim={5.5cm 1.6cm 7.2cm 5cm},clip]{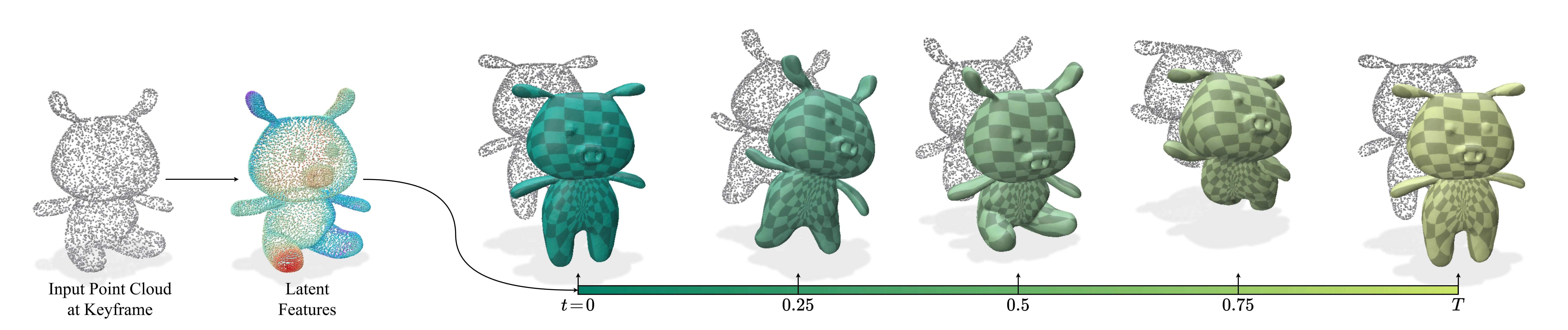} 
\captionof{figure}{We present Neu-PiG, a method that learns spatially smooth and temporally coherent deformations from dynamic point clouds. Starting from input point clouds (left), a unified latent space parameterized by an initial reference mesh (middle) is optimized through multi-scale grids to produce high-fidelity deformations (right) without relying on strong priors or correspondences. \vspace{1em}}
\label{fig:teaser}
}]

\input{sec/0_abstract}
\input{sec/1_introduction}
\input{sec/2_related_work}

\input{sec/3_method}
\input{sec/4_experiments}
\input{sec/5_conclusion}
{
    \small
    \bibliographystyle{ieeenat_fullname}
    \bibliography{main}
}

\input{sec/X_suppl}

\end{document}

%% file: preamble.tex
\usepackage{comment}
\usepackage{xspace}

\newcommand{\todo}[1]{{\color{red}#1}}
\newcommand{\TODO}[1]{\textbf{\color{red}[TODO: #1]}}

\usepackage{pifont}

\usepackage{multirow}
\usepackage{mwe} 
\usepackage{subcaption} 
\usepackage[normalem]{ulem}

\newcommand{\abs}[1]{\left\vert#1\right\vert}
\newcommand{\norm}[1]{\left\Vert#1\right\Vert}

\usepackage{bm}

\renewcommand{\vec}[1]{\bm{#1}}
\newcommand{\mat}[1]{\bm{#1}}
\newcommand{\set}[1]{\mathcal{#1}}

\DeclareMathOperator*{\argmax}{arg\,max}
\DeclareMathOperator*{\sg}{sg}

\let\originalleft\left
\let\originalright\right
\renewcommand{\left}{\mathopen{}\mathclose\bgroup\originalleft}
\renewcommand{\right}{\aftergroup\egroup\originalright}



%% file: sec/0_abstract.tex
\begin{abstract}
Temporally consistent surface reconstruction of dynamic 3D objects from unstructured point cloud data remains challenging, especially for very long sequences.
Existing methods either optimize deformations incrementally, risking drift and requiring long runtimes, or rely on complex learned models that demand category-specific training.
We present \emph{Neu-PiG}, a fast deformation optimization method based on a novel preconditioned latent-grid encoding that distributes spatial features parameterized on the position and normal direction of a keyframe surface.
Our method encodes entire deformations across all time steps at various spatial scales into a multi-resolution latent grid, parameterized by the position and normal direction of a reference surface from a single keyframe.
This latent representation is then augmented for time modulation and decoded into per-frame 6-DoF deformations via a lightweight multi-layer perceptron (MLP).
To achieve high-fidelity, drift-free surface reconstructions in seconds, we employ Sobolev preconditioning during gradient-based training of the latent space, completely avoiding the need for any explicit correspondences or further priors.
Experiments across diverse human and animal datasets demonstrate that Neu-PiG outperforms state-the-art approaches, offering both superior accuracy and scalability to long sequences while running at least 60$\times$ faster than existing training-free methods and achieving inference speeds on the same order as heavy pre-trained models.
\end{abstract}

%% file: sec/1_introduction.tex
\section{Introduction}
\label{sec:intro}

Understanding and modeling how shapes in the surrounding world evolve over time is fundamental to perceiving and interacting with dynamic environments.
Whether observing a human performing complex motions, an animal moving through its habitat, or an object undergoing physical transformations, capturing such spatiotemporal deformations is key to reasoning about the physical and semantic structure of the world.
This capability underpins a wide range of applications in augmented and virtual reality (AR/VR), robotics, autonomous systems, and computer vision, including realistic motion capture, scene reconstruction, and human–robot interaction. 

A common strategy to represent deformable shapes is through parametric models, which provide a controllable yet expressive space of possible deformations. Prominent examples include SMPL~\cite{loper2015smpl} and SMPL-X~\cite{pavlakos2019expressive} for human bodies, FLAME~\cite{li2017learning} for facial geometry, and MANO~\cite{romero2017embodied} for hands.
While these models offer compact and interpretable control, they are inherently limited to the object categories for which they were designed.
To move beyond category-specific representations, template-free methods have been proposed, which can broadly be divided into optimization-based and learning-based approaches. Optimization-based methods (\eg, DynoSurf~\cite{yao2024dynosurf}, PDG~\cite{kaltheuner2025preconditioned}) directly optimize surface deformations for each sequence, producing high-quality results but at the cost of long runtimes.
In contrast, learning-based approaches amortize reconstruction through category-specific priors, achieving fast inference but struggling to generalize beyond their training domain.
Despite their complementary strengths, a general, efficient, and category-agnostic approach for modeling dynamic shape deformations remains an open challenge. 

To address these limitations, we propose \emph{Neu-PiG}, a novel method which introduces \textbf{Neu}ral \textbf{P}recond\textbf{i}tioned \textbf{G}rids as an efficient representation for fast estimation of spatially smooth and temporally coherent deformations from dynamic point cloud sequences.
Starting from an initial mesh estimate as a canonical reference, we learn the complete deformation across all time steps via a novel surface encoding that is parameterized by vertex positions and normal directions.
%
To enable this without relying on correspondences or restricting the method's generalizability by strong priors, our key insight lies in the geometric structure and the associated optimization procedure of the latent embedding.
We store learnable feature vectors at different spatial scales in a multi-resolution voxel grid and aggregate them for a given query point via trilinear interpolation and averaging.
This ensures that we learn the absolute deformations at each level rather than a decomposition of them which would be sensitive to small deviations and thus significantly less stable during optimization.
We further stabilize this process by applying Sobolev preconditioning to the gradient-based training of the latent grid.
Consequently, this approach allows us to obtain a unified latent space which can be augmented for time-modulation and then decoded into high-fidelity deformations via a lightweight MLP.
As illustrated in \cref{fig:teaser} and demonstrated in extensive experiments, Neu-PiG achieves superior reconstruction accuracy at an order-of-magnitude lower runtimes than state-of-the-art optimization-based approaches and comparable to the inference times of pretrained models.

In summary, our key contributions are: 
\begin{itemize}
    \item We propose a fast optimization-based method that estimates temporally coherent deformations of arbitrary subjects, including humans and animals, from sequential point cloud data.
    \item We introduce a novel preconditioned surface encoding based on vertex positions and normal directions of a reference mesh that captures the full deformations across all time steps into a unified latent space.
    \item We design a multi-scale latent grid representation to enforce spatial consistency both on global and local scales, enabling fast decoding via a lightweight MLP.
\end{itemize}
Code is available at: \href{https://github.com/vc-bonn/neu-pig}{https://github.com/vc-bonn/neu-pig}

%% file: sec/2_related_work.tex
\section{Related Work}
\label{sec:relatedwork}

\subsection{Parametric Template Models}

Category-specific reconstruction has historically relied on deforming a canonical mesh to match observations~\cite{li2017learning, wang2022faceverse, loper2015smpl, pavlakos2019expressive, anguelov2005scape, xu2020ghum}.  
Such models represent facial and body geometry within learned low-dimensional manifolds, enabling expressive control of identity and pose~\cite{bogo2016keep, sanyal2019learning, liu2021neural}.  
Extensions further incorporate clothing, soft tissue, or body-shape variability~\cite{bhatnagar2019multi, alldieck2019learning, tiwari2020sizer}, making parametric systems dominant for structured capture tasks.
Despite their success, the linear subspace assumption of these models restricts generalization beyond the training domain~\cite{loper2015smpl, xu2020ghum}.
Recent neural formulations replace explicit templates with implicit shape functions that learn deformation spaces directly from data~\cite{li2017learning, wang2022faceverse}.  
Current state-of-the-art methods decouple skeletal bases from surface-shape representations~\cite{park2025atlas}, or body-shape from clothing \cite{chen2025d}, while complementary approaches bridge parametric and implicit formulations by learning MLP weights for an SDF-based body model~\cite{mihajlovic2025volumetricsmpl}.
 


Our approach follows this trend toward implicit representations but removes category dependence entirely: \mbox{Neu-PiG} reconstructs dynamic surfaces from unstructured sequences using a latent spatial grid and a lightweight decoder without predefined templates.

\begin{figure*}
    \centering
        \includegraphics[width=0.92\textwidth, trim={1.6cm 3.0cm 1.8cm 4.1cm},clip]{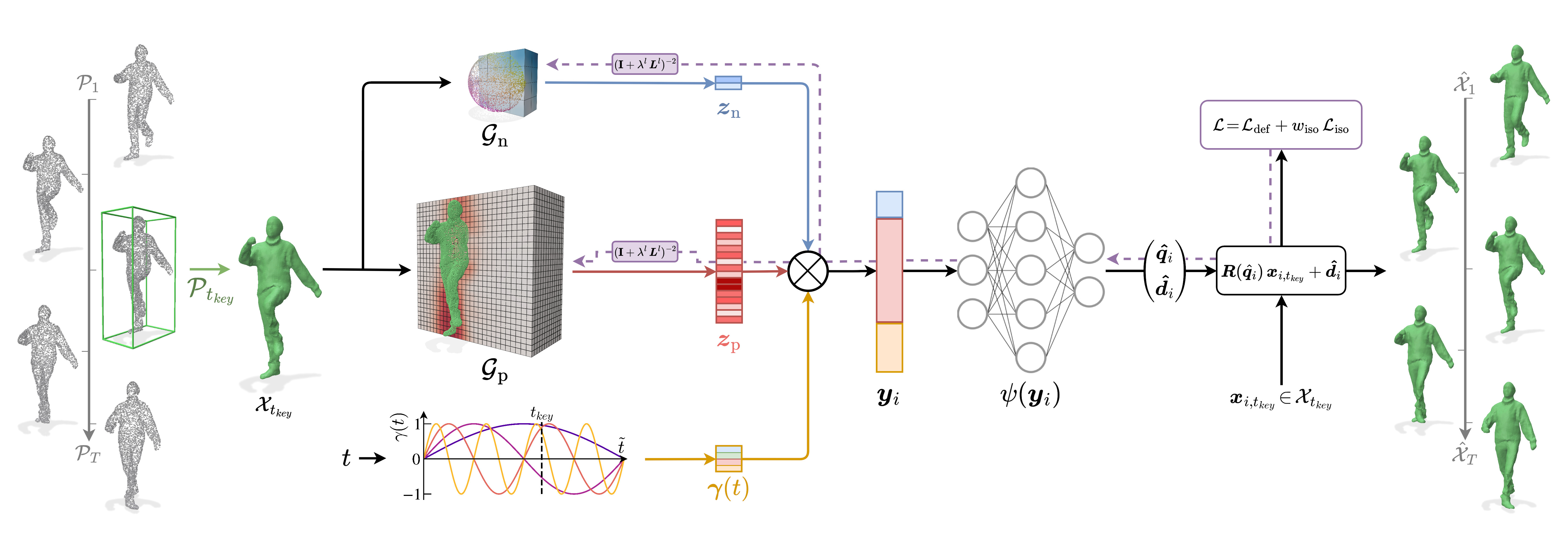} 
        \caption{
        Overview of Neu-PiG.
        A reference surface $\set{X}_{t_{\mathrm{key}}}$ is first generated from the input point cloud at a keyframe $\set{P}_{t_{\mathrm{key}}}$.
        Position and normal-direction-embedded latent features $\vec{z}_{\mathrm{p}}$ and $\vec{z}_{\mathrm{n}}$ are then sampled from multi-resolution preconditioned grids and combined with a time embedding $\vec{\gamma}(t)$.
        A lightweight MLP $\psi$ predicts per-frame 6-DoF deformations that are applied to $\set{X}_{t_{\mathrm{key}}}$ to reconstruct each frame $ \hat{\set{X}}_t$.
        Sobolev preconditioning in the latent space enforces spatial smoothness and temporal coherence across the sequence during optimization.
        }
        \label{fig:overview}
\end{figure*}

\subsection{Deformation Field Estimation}

Non-rigid motion recovery aims to describe how points or surfaces evolve across time~\cite{sumner2007embedded, newcombe2015dynamicfusion}.  
Early methods estimated dense warp fields via optimization, \eg through embedded graphs~\cite{sumner2007embedded} or volumetric tracking~\cite{newcombe2015dynamicfusion}.  
Subsequent dynamic radiance formulations employ Gaussian primitives for compact 4D mapping~\cite{kerbl20233d, matsuki20254dtam} or separate static and moving components for SLAM-style pipelines~\cite{li2025dynagslamrealtimegaussiansplattingslam, zheng2025wildgsslammonoculargaussiansplatting}.  

Neural deformation models shifted the focus from explicit alignment to learned motion inference~\cite{bozic2021neural, bozic2020neural, li2022non}.  
Hierarchical and continuous formulations parameterize motion as neural vector fields~\cite{niemeyer2019occupancy, chen2018neural, jiang2021learning, huang2020meshode, tang2021learning} or as spatio-temporal surfaces~\cite{nizamani2025dynamicneuralsurfaceselastic}.  
Probabilistic variants treat deformation as a distribution over 4D geometry~\cite{cao2024motion2vecsets}, while canonical-space techniques couple shape and motion within rendering frameworks~\cite{pumarola2021d, park2021nerfies, li2021neural, lei2022cadex, yenamandra2021i3dmm}.  
Optimization-only strategies such as DynoSurf~\cite{yao2024dynosurf} achieve supervision-free tracking but remain computationally demanding.

\paragraph{Latent and Implicit Representations.}
Recent advances move beyond explicit displacement fields by storing deformation cues in spatially distributed latent representations that are decoded into motion~\cite{pumarola2021d, park2021nerfies}. 
Dense voxel features paired with compact decoders provide locality but can be memory-intensive at high resolutions~\cite{li2022non, yao2024dynosurf}. 
As a foundational alternative for neural fields, multi-resolution hash encodings supply compact capacity for static reconstruction~\cite{muller2022instant}; however, their collision-prone, non-smooth parameterization is ill-suited to coherent non-rigid motion and, to our knowledge, has not been adopted for deformation modeling. 
For dynamic scenes, factorized latent structures impose stronger continuity priors across space and time-tensor decompositions and separable planes yield smoother fields with lightweight decoders~\cite{chen2022tensorf, fridovich2023k, cao2023hexplane}. 
In parallel, implicit deformation fields map canonical coordinates to observations via learned warps, often combining latent features with time embeddings for scalable long-horizon tracking~\cite{pumarola2021d, park2021nerfies, cao2024motion2vecsets}. 
Complementary trends instantiate related ideas in alternative primitives by embedding motion in per-Gaussian codes for deformable 3D splats~\cite{bae2024per}. 
Following these directions, but focusing specifically on stable latent carriers for deformation, we group our design around a preconditioned voxel representation decoded by a lightweight MLP; for broader context on latent field formulations in dynamic settings, see~\cite{zhu2025dynamic}.

Neu-PiG differs from explicit per timestep deformation grid formulations such as Preconditioned Deformation Grids~\cite{kaltheuner2025preconditioned} by maintaining a single preconditioned latent grid shared across all frames, yielding smooth, drift-free results on extended sequences.

\subsection{Preconditioning}

Stabilizing gradient-based optimization through preconditioning is well-established in numerical geometry~\cite{claici2017isometry, kovalsky2016accelerated, krishnan2013efficient}, and while challenging in high-dimensional latent spaces, graph- and learning-based variants adapt such schemes for data-driven problems~\cite{rudikov2024neural, li2023learning, chen2024graph, hausner2024neural, trifonov2024learning}.

Sobolev gradient methods~\cite{neuberger1985steepest} compute updates under smoother inner products, improving convergence in shape and surface optimization~\cite{renka2004constructing, renka1995minimal, eckstein2007generalized, martin2013efficient, kaltheuner2025preconditioned}, and have shown benefits in training scenarios \cite{cho2025sobolev, oh2025sobolev, kilicsoy2024sobolev, baravdish2024hybrid}, \eg by accelerating  ReLU network convergence~\cite{oh2025sobolev}. %
They have since appeared in non-rigid reconstruction~\cite{slavcheva2018sobolevfusion, jung2025preconditioned}, differentiable rendering~\cite{nicolet2021large}, and spatiotemporal filtering~\cite{chang2024spatiotemporal}.  

Neu-PiG extends this principle to latent voxel spaces, propagating Sobolev-filtered gradients through a learned feature grid to ensure smooth, coherent updates of latent codes across long dynamic sequences.

%% file: sec/3_method.tex
\section{Method}
\label{sec:method}
Given a sequence of unstructured point clouds $ \{ \set{P}_t\}_{t = 1}^T $ where each $ { \set{P}_t = \{ \vec{p}_{i,t} \in \mathbb{R}^3 \}_{i = 1}^{N_t} } $, we address the problem of dynamic surface reconstruction by estimating deformations from a single reference mesh defined at a keyframe $ { t_\mathrm{key} \in \{1, \dots, T\} } $, to all other frames.
An overview of this setup is illustrated in \cref{fig:overview}.
The reference mesh consists of vertices and their associated vertex normals, $ { \set{X}_{t_\mathrm{key}} = \{ (\vec{x}_{i,t_\mathrm{key}}, \vec{n}_{i,t_\mathrm{key}}) \in \mathbb{R}^3 \times \mathbb{R}^3 \}_{i = 1}^{N_{t_\mathrm{key}}} } $ (see \cref{sec:initialization}).
We encode the deformations into smooth latent features based on the vertex positions and normal directions of $\set{X}_{t_\mathrm{key}}$ and store them in a pair of preconditioned multi-resolution voxel grids. 
Our latent-grid formulation can be interpreted as a spatially coherent spatio-temporal field optimized under Sobolev regularization.
Unlike implicit neural representations that encode geometry entirely in network weights, our factorized grid–decoder formulation distributes high-dimensional features over space, enabling a MLP $\psi$ to reconstruct dynamic geometry from local content.
This interpretation motivates our preconditioned latent-field design described in \cref{sec:latent_grid}.
These latent encodings are combined with frequency-encoded timesteps $ t $ and decoded by a lightweight MLP $\psi$ that predicts per-frame deformations, transforming the reference mesh $\set{X}_{t_\mathrm{key}}$ into temporally coherent reconstructions $ \{ \hat{\set{X}}_t \}_{t = 1}^T $, see \cref{sec:temporal_deformation_model,sec:transformation_mapping}.
We jointly optimize the voxel grids and network parameters using a deformation loss and an isometry loss, without any additional regularization or priors, see \cref{sec:optimization_objectives}.

\subsection{Initialization}
\label{sec:initialization}
We begin by selecting a keyframe $ t_\mathrm{key} $, from which we can reconstruct a surface mesh with suitable topology. 
This frame acts as a canonical reference for all subsequent deformations.
Following the method of PDG~\cite{kaltheuner2025preconditioned}, we assess the spatial extent of each input point cloud $ \set{P}_t $, weighted by a bias term that favors the temporal midpoint of the sequence.
The point cloud corresponding to the selected keyframe, $ \set{P}_{t_{\mathrm{key}}} $, is then converted into an initial mesh $ \set{X}_{t_{\mathrm{key}}} $ using screened Poisson surface reconstruction~\cite{kazhdan2013screened}.
Note that the precise geometric alignment of $ \set{X}_{t_{\mathrm{key}}} $ with $ \set{P}_{t_{\mathrm{key}}} $ is not critical, as long as the topology is correct, since we estimate deformations for all frames, including $ t_{\mathrm{key}} $.
%

We initialize the latent features stored at each level of the multi-resolution voxel grid with zero vectors.
Furthermore, we set the weights of all but the last layer of the MLP to random values, while zeroing the weights of the last one.
This initializes the deformations as identity transformations and lets them evolve gradually during optimization, avoiding sudden jumps that could disrupt latent-space consistency and lead to incoherent results.

\subsection{Multi-Resolution Latent Grids}
\label{sec:latent_grid}

To efficiently represent the geometric structure across spatial scales, we encode the reference surface $\set{X}_{t_{\mathrm{key}}}$ in a hierarchy of preconditioned voxel grids.
Each grid level stores latent features at a distinct spatial frequency: coarser levels capture global deformation patterns, while finer levels encode high-frequency geometric detail.
%
%
This hierarchical representation serves as an expressive latent structure for the deformation network introduced in \cref{sec:temporal_deformation_model}.

\paragraph{Position and Normal-Direction Embeddings.}
For each vertex $ { \vec{v}_{i,t_\mathrm{key}} = (\vec{x}_{i,t_\mathrm{key}}, \vec{n}_{i,t_\mathrm{key}}) } $ on the reference surface, we query two distinct grids: a position-based grid $\set{G}_{\mathrm{p}}$ that encodes spatial context, and a normal-direction-based grid $\set{G}_{\mathrm{n}}$ that captures local orientation information.
This enables spatially adjacent regions with differing normal directions to deform independently.
Each grid cell stores a learnable feature vector and we retrieve interpolated features at $ \vec{x}_{i,t_\mathrm{key}} $ and $ \vec{n}_{i,t_\mathrm{key}} $ via trilinear interpolation, yielding the latent vectors $ \vec{z}_{\mathrm{p}}(\vec{x}_{i,t_\mathrm{key}}) $ and $ \vec{z}_{\mathrm{n}}(\vec{n}_{i,t_\mathrm{key}}) $, respectively.
As the position-based embedding is the most important factor in deformation, we store 30-dimensional latent features at each cell of $\set{G}_{\mathrm{p}}$, resulting in $ { \vec{z}_{\mathrm{p}} \in \mathbb{R}^{30} } $.
In contrast, the orientation-based embedding in $\set{G}_{\mathrm{n}}$ stores 2-dimensional latent features at each grid cell, resulting in $ { \vec{z}_{\mathrm{n}} \in \mathbb{R}^2 } $.

\paragraph{Grid Hierarchy.}
The position grid $\set{G}_{\mathrm{p}}$ is defined as a multi-resolution hierarchy with progressively increasing spatial resolution~\cite{kaltheuner2025preconditioned}.
It consists of eight levels, with the coarsest grid containing $2^3$ cells and the finest $32^3$ cells, achieved by increasing the grid resolution by 3 elements in each level.
Instead of treating the grid levels independently, we aggregate their outputs into a unified latent representation, ensuring that each point $ \vec{x}_{i,t_\mathrm{key}} $ is associated with a spatially coherent latent feature:
\begin{equation}
    \vec{z}_{\mathrm{p}}(\vec{x}_{i,t_\mathrm{key}}) = \dfrac{1}{L} \sum_{l = 1}^{L} \vec{z}_{\mathrm{p}}^l(\vec{x}_{i,t_\mathrm{key}}).
\end{equation}
The normal-direction-based grid $\set{G}_{\mathrm{n}}$ is defined as a single-resolution grid of size $4^3$, which is sufficient to encode local orientation changes, despite the potentially large normal variations in nearby regions.

\subsection{Temporal Deformation Model}
\label{sec:temporal_deformation_model}
Our temporal deformation model predicts per-frame transformations that deform the reference mesh $\set{X}_{t_{\mathrm{key}}}$ into temporally varying reconstructions $ \{ \hat{\set{X}}_t \}_{t = 1}^T $.
Each vertex in $\set{X}_{t_{\mathrm{key}}}$ is represented by two latent descriptors from the preconditioned voxel grids $\set{G}_{\mathrm{p}}$ and $\set{G}_{\mathrm{n}}$, a position-based encoding $\vec{z}_{\mathrm{p}}(\vec{x}_{i,t_{\mathrm{key}}})$ and a normal direction-based encoding $\vec{z}_{\mathrm{n}}(\vec{n}_{i,t_{\mathrm{key}}})$.
For each timestep $t$, these descriptors are combined with a time embedding $\vec{\gamma}(t)$ to form the input vector:
\begin{equation}
    \vec{y}_i = \left( \vec{z}_{\mathrm{n}}(\vec{n}_{i,t_{\mathrm{key}}}), \vec{z}_{\mathrm{p}}(\vec{x}_{i,t_{\mathrm{key}}}), \vec{\gamma}(t) \right)^T \in \mathbb{R}^{2 + 30 + 8}.
\end{equation}
This input formulation provides the network with both spatial and directional context, enabling the generation of temporally coherent deformations that are locally smooth and globally consistent across the sequence.

\paragraph{Fourier Time Encoding.}
To represent time in a continuous, multi-scale manner, we normalize the timestep to the range $ [0, 1] $ via $ { \tilde{t} = (t - 1) / (T - 1) } $ and map it to a sinusoidal embedding following the Fourier feature formulation~\cite{tancik2020fourier}.
Specifically, we define the time embedding as:
\begin{equation}
    \vec{\gamma}(t) = \left[ \sin(\pi \nu_j \tilde{t}), \cos(\pi \nu_j \tilde{t}) \right]_{j=1}^{M} \in \mathbb{R}^{2M},
\end{equation}
where each frequency $ { \nu_j = 2^{j - 1} } $ modulates the temporal signal at a distinct scale.
This encoding allows the network to capture both low-frequency, slowly varying motions and high-frequency, transient deformations.
In practice, we use $ { M = 4 } $ frequencies, resulting in an 8D time embedding.

\paragraph{Network Architecture.}
The deformation network $\psi$ is implemented as a lightweight MLP that maps the input vector $\vec{y}$ to a transformation consisting of a rotation quaternion $ { \vec{q} \in \mathbb{R}^4 } $ and a displacement vector $ { \vec{d} \in \mathbb{R}^3 } $:
\begin{equation}
    (\vec{q}, \vec{d})^T = \psi(\vec{y}).
\end{equation}
The network comprises three fully connected layers, each with 512 hidden units and LeakyReLU activations.
The final layer outputs a 7-dimensional transformation vector.
All parameters of $\psi$ are optimized jointly with the latent grid features during the reconstruction process.

\subsection{Transformation Mapping}
\label{sec:transformation_mapping}
To apply the predictions from the network $\psi$ while ensuring stability and smoothness during optimization, we map each component to explicit geometric transformations.

\paragraph{Rotation Component.}
Let $ { \vec{q} = (q_w, q_x, q_y, q_z)^T \in \mathbb{R}^4 } $ denote the quaternion parameters predicted by the deformation network.
To enforce that a zero-valued network output corresponds to the identity rotation, we offset the scalar component $q_w$ by $1$, i.e.,
\begin{equation}
    \hat{\vec{q}} = \frac{(1 + q_w, q_x, q_y, q_z)^T}{\norm{(1 + q_w, q_x, q_y, q_z)^T}},
\end{equation}
and subsequently normalize the quaternion to unit length.
This formulation provides a continuous, stable rotation representation, avoiding singularities in \eg Euler angles.

\paragraph{Translation Component.}
To decouple rotational and displacement effects, the network predicts a displacement vector $ { \vec{d} \in \mathbb{R}^3 } $.
Since all points lie within the normalized coordinate space $[-1, 1]^3$, we constrain displacements by applying a hyperbolic tangent function:
\begin{equation}
    \hat{\vec{d}} = \tanh(\alpha \, \vec{d}), \qquad \alpha = 0.1.
\end{equation}
This formulation prevents excessive displacements while still allowing smooth, learnable displacements.

\paragraph{Final Transformation.}
The complete transformation applied to each vertex $ { \vec{x}_{i,t_{\mathrm{key}}} \in \set{X}_{t_{\mathrm{key}}} } $ can now be expressed as:
\begin{equation}
    \hat{\vec{x}}_{i,t} = \mat{R}(\hat{\vec{q}}_i) \, \vec{x}_{i,t_{\mathrm{key}}} + \hat{\vec{d}}_i.
\end{equation}
Here, $ \mat{R}(\hat{\vec{q}}) $ denotes the rotation matrix corresponding to the unit quaternion $ \hat{\vec{q}} $.
Together, the per-vertex quaternion-based rotations and bounded displacements form a compact and stable transformation representation, facilitating smooth temporal deformations across the sequence.

\subsection{Optimization Objectives}
\label{sec:optimization_objectives}
Starting from the reference surface $\set{X}_{t_\mathrm{key}}$ defining the object topology, we optimize deformation fields that map it directly to each target point cloud $\set{P}_t$.
Our objective combines a deformation loss and an isometry loss, which together balance reconstruction accuracy and surface smoothness:
\begin{equation}
    \mathcal{L} = \mathcal{L}_{\mathrm{def}} + w_{\mathrm{iso}} \, \mathcal{L}_{\mathrm{iso}}.
\end{equation}
As our MLP architecture inherently promotes smooth deformations, we use a relatively small isometry weight $ { w_{\mathrm{iso}} = 100 } $ which downweights $ \mathcal{L}_{\mathrm{iso}} $ to approximately 10\% of the magnitude of $ \mathcal{L}_{\mathrm{def}} $.

\paragraph{Preconditioning.}
To improve convergence stability, we follow PDG~\cite{kaltheuner2025preconditioned} and apply Sobolev preconditioning~\cite{nicolet2021large} to smooth gradient updates of the latent grids $\set{G}_{\mathrm{p}}$ and $\set{G}_{\mathrm{n}}$.
The parameters $\vec{z}^l$ of grid level $l$ are updated as
\begin{equation}
	\vec{z}^l \leftarrow \vec{z}^l - \eta \, (\mathbf{I} + \lambda^l \, \mat{L}^l)^{-2} \, \frac{\partial \set{L}}{\partial \vec{z}^l},
\end{equation}
where $ \eta $ denotes the learning rate, $ \mathbf{I} $ is the identity matrix, $ { \lambda^l > 0 } $ controls the strength of spatial smoothing, and $ \mat{L}^l $ is the Laplacian matrix at level $ l $.
%
%
Here, $ { (\mathbf{I} + \lambda^l \, \mat{L}^l)^{-2} } $ acts as a low-pass filter on the gradient, coupling neighboring cells and promoting spatially coherent latent updates that remain temporally stable across the sequence.
The bandwidth parameter $\lambda^l$ can be tuned to match the decoder's effective frequency capacity~\cite{xu2022signal,shi2024fourier}, ensuring that latent-field variations remain within the MLP’s learnable range.
In contrast to PDG~\cite{kaltheuner2025preconditioned}, which preconditions raw deformation fields, Neu-PiG applies Sobolev preconditioning to learned latent vectors, allowing high-dimensional features to capture richer local variations, while a single MLP ensures temporally consistent behavior across space and time.
%

\begin{figure*}
    \centering
    \includegraphics[width=0.92\linewidth, trim={0cm 3.2cm 2.2cm 0cm},clip]{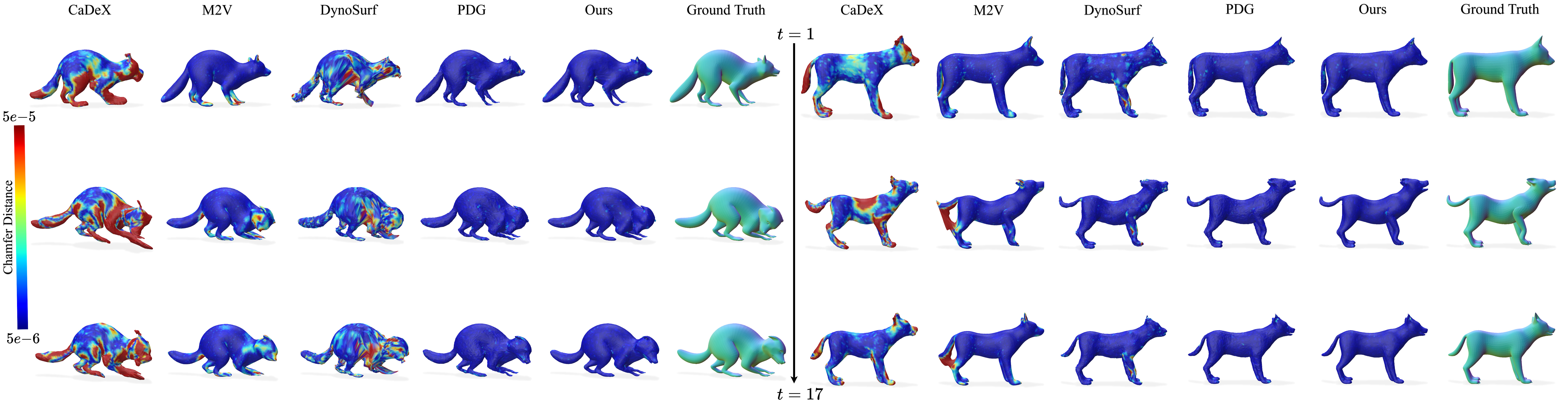}
    \caption{Qualitative reconstructions on DT4D. Neu-PiG preserves geometry and temporal consistency across complex animal motions.}
    \label{fig:dt4d}
\end{figure*}

\paragraph{Deformation Loss.}
The main objective function measures the alignment between the deformed surface $\hat{\set{X}}_t$ and the input point cloud $\set{P}_t$ at each timestep:
\begin{equation}
    \mathcal{L}_{\mathrm{def}} = \frac{1}{T} \sum_{t = 1}^{T} w_{\mathrm{conf}}(t) \cdot L_{\mathrm{CD}}(\hat{\set{X}}_t, \set{P}_t),
\end{equation}
where $L_{\mathrm{CD}}$ is the robust Chamfer distance~\cite{yao2024dynosurf}, and $w_{\mathrm{conf}}(t)$ a time-adaptive confidence weight, originally proposed in PDG~\cite{kaltheuner2025preconditioned}.
Although our method estimates the total deformation from the reference surface rather than deformation between consecutive frames, confidence weighting is beneficial, particularly when combined with our frequency-split timestep encoding, which implicitly accounts for past reconstruction accuracy.
We define the weight as
\begin{gather}
    w_{\mathrm{conf}}(t) = \prod_{\tau = t_{\mathrm{key}}}^{t} \omega(\tau)^{\delta},
    \label{eq:confidence}
    \\
    \omega(\tau) = \frac{1}{1 + \max(0, \mathrm{cd}_{\tau} - \mathrm{cd}_{t_{\mathrm{key}}})} \in [0, 1],
\end{gather}
where $\omega(\tau)$ represents the reconstruction performance at time step $\tau$ relative to the keyframe $ t_{\mathrm{key}} $.
We compute this weight via the Chamfer distance $ { \mathrm{cd}_{\tau} = L_{\mathrm{CD}}(\sg(\hat{\set{X}}_{\tau}), \set{P}_{\tau}) } $, where $ \sg $ denotes the stop-gradient operator.
Note that $ { \omega(t_{\mathrm{key}}) = 1 } $.
Additionally, to mitigate optimization deadlocks and improve convergence stability, we employ a catch-up variable $ { \delta = 1 - \sqrt{\bar{e}} } $, which gradually increases the confidence during training.
Here, $ { \bar{e} = e / e_{\mathrm{max}} } $ denotes the normalized optimization epoch.

\paragraph{Isometry Loss.}
To preserve local structure, we use an isometry loss that penalizes changes in edge lengths $ { \hat{\vec{e}}_{ij,t} = \hat{\vec{x}}_{i, t} - \hat{\vec{x}}_{j, t} } $ of the deformed meshes:
\begin{equation}
    \mathcal{L}_{\mathrm{iso}} = \frac{1}{T \, \abs{\set{E}}} \sum_{t = 1}^T \sum_{(i,j) \in \set{E}} \Big\vert \norm{\hat{\vec{e}}_{ij, t}} - \sg\left(\norm{\hat{\vec{e}}_{ij, t_{\mathrm{key}}}}\right) \Big\vert,
\end{equation}
where $\set{E}$ denotes the set of mesh edges.

\paragraph{Implementation Details.}
We jointly optimize the grid features and MLP parameters using the Adam optimizer~\cite{kingma2015adam} with its default parameters. 
The MLP uses a learning rate of $10^{-3}$, while each grid level $l$ follows a per-level rate of $0.005\times2.5^l$. 
The smoothness weight is defined as $\lambda = 0.4\times1.5^l$. 
To highlight the trade-off between accuracy and runtime, we report two configurations evaluated on a NVIDIA RTX 4090 that differ only in optimization length, 250 epochs (Ours\textsuperscript{$\dagger$}) and 1000 epochs (Ours).

%% file: sec/4_experiments.tex
\section{Evaluation}
\label{sec:evaluation}

\begin{table}
    \scriptsize
    \centering
    \caption{Quantitative comparison on DFAUST, AMA, and DT4D. Neu-PiG achieves best accuracy and temporal coherence while being 60$\times$ faster than prior training-free methods.}
    \renewcommand{\arraystretch}{1.1}
    \setlength{\tabcolsep}{4.5pt}
    \begin{tabular}{l|c|ccccc}
        \toprule
        \multicolumn{1}{c}{} & & CD $ [\times 10^{-5} ] $ $\downarrow$ & NC $\uparrow$ & F-$0.5\%$ $\uparrow$  & Corr. $\downarrow$ & Time $\downarrow$\\
    
        \midrule
        \multirow{6}{*}{\rotatebox{90}{\tiny DFAUST}}
        & CaDeX & \phantom{0}3.68 & 0.941 & 0.730 & 0.013 & \textbf{\phantom{0}3\,\text{s}\phantom{mi}} \\
        & DynoSurf & \phantom{0}2.13 & 0.953 & 0.980 & 0.010 & 30\,\text{min} \\
        & M2V & \phantom{0}1.61 &  0.960 & 0.877 & - & \phantom{0}4\,\text{s}\phantom{mi} \\
        & PDG & \phantom{0}0.46 & 0.956 & 0.987 & 0.017 & \phantom{0}7\,\text{min}  \\
        & Ours\textsuperscript{$\dagger$}  & \phantom{0}0.40 & 0.967 & \textbf{0.989} & \textbf{0.008} & \phantom{0}8\,\text{s}\phantom{mi} \\
        & Ours\phantom{0} & \phantom{0}\textbf{0.39} & \textbf{0.968} & \textbf{0.989} & 0.009 & 32\,\text{s}\phantom{mi} \\
        
        \midrule
        \multirow{6}{*}{\rotatebox{90}{\tiny DT4D}}
        & CaDeX & 56.51 & 0.870 & 0.386 & 0.050 & \textbf{\phantom{0}3\,\text{s}\phantom{mi}} \\
        & DynoSurf & 15.18 & 0.919 & 0.773 & \textbf{0.032} & 30\,\text{min} \\
        & M2V & \phantom{0}7.61 & 0.944 & 0.792 & - & \phantom{0}4\,\text{s}\phantom{mi} \\
        & PDG & \phantom{0}1.67 & 0.956 & 0.948 & 0.043 & \phantom{0}7\,\text{min}\\
        & Ours\textsuperscript{$\dagger$} & \phantom{0}0.96 & \textbf{0.969} & 0.962 & 0.034 & \phantom{0}8\,\text{s}\phantom{mi} \\
        & Ours\phantom{0} & \phantom{0}\textbf{0.87} & \textbf{0.969} & \textbf{0.968} & 0.034 & 32\,\text{s}\phantom{mi}\\ 
        
        \midrule
        \multirow{4}{*}{\rotatebox{90}{\tiny AMA}}
        & DynoSurf & \phantom{0}1.01 & 0.918 & 0.921 & 0.044 & 30\,\text{min} \\
        & PDG & \phantom{0}0.79 & 0.926 & 0.961 & 0.037 & \phantom{0}7\,\text{min} \\
        & Ours\textsuperscript{$\dagger$} & \phantom{0}0.53 & 0.946 & 0.978 & 0.024 & \textbf{\phantom{0}8\,\text{s}\phantom{mi}} \\
        & Ours\phantom{0} & \phantom{0}\textbf{0.44} & \textbf{0.951} &\textbf{0.988} & \textbf{0.018} & 32\,\text{s}\phantom{mi} \\
        \bottomrule
    \end{tabular}
    \label{tab:metrics}
\end{table}

\begin{figure}
    \centering
    \includegraphics[width=0.92\linewidth, trim={0cm 0.25cm 0cm 0.1cm},clip]{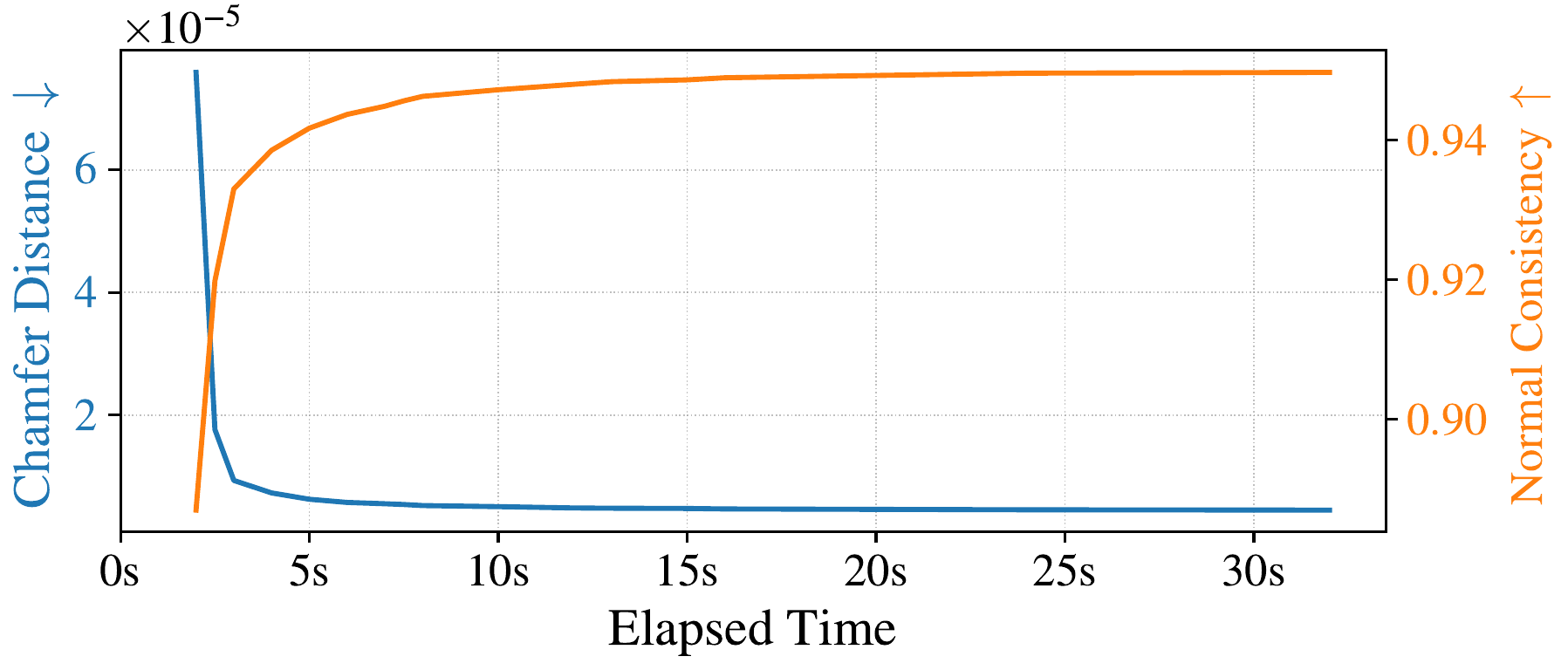}
    \caption{Convergence of Chamfer Distance and Normal Consistency on DT4D, showing high-quality reconstructions in seconds.}
    \label{fig:reconstruction_length}
\end{figure}

We evaluate Neu-PiG against state-of-the-art methods, distinguishing between learning-based approaches that rely on category-specific priors and training-free methods that directly optimize deformations from point clouds without pretraining. 
The learning-based baselines include CaDeX~\cite{lei2022cadex} and M2V~\cite{cao2024motion2vecsets}, the latter requiring predefined inter-frame correspondences.
For training-free baselines, we compare against DynoSurf~\cite{yao2024dynosurf} and PDG~\cite{kaltheuner2025preconditioned}. 
Experiments are conducted on three public benchmarks: DFAUST~\cite{dfaust} (human motion), AMA~\cite{ama} (clothed human performance), and DT4D~\cite{dt4d} (articulated animals). 
Unless stated otherwise, each sequence contains ${T = 17}$ frames following previous works~\cite{kaltheuner2025preconditioned,yao2024dynosurf} and consists of 5000 points per timestep.

\begin{figure*}
    \centering
    \includegraphics[width=0.92\linewidth, trim={0cm 2cm 0cm 0.8cm},clip]{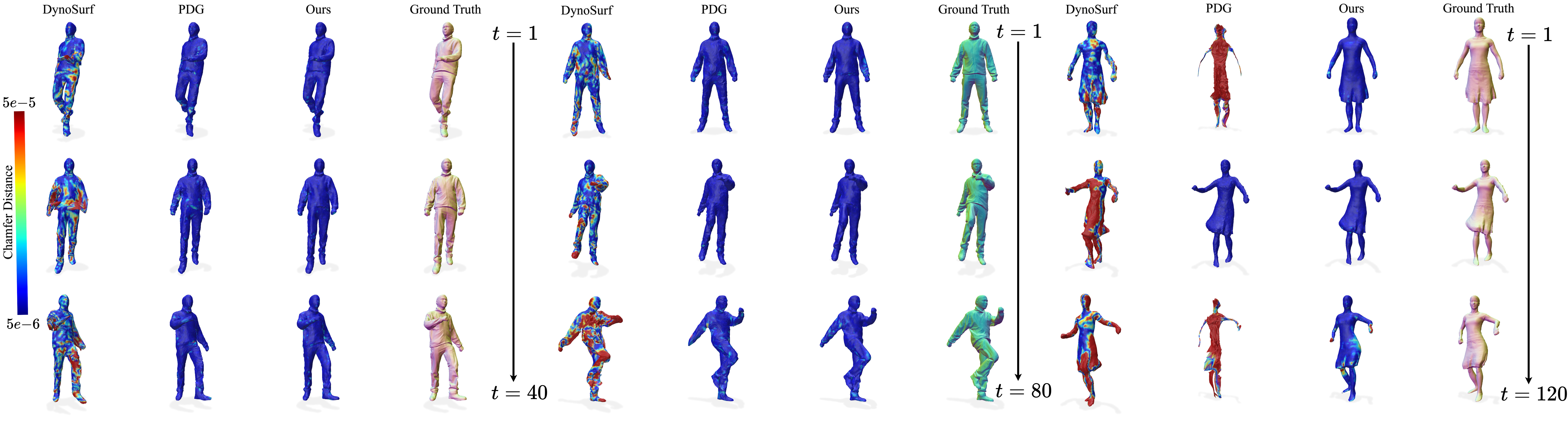}
    \caption{Performance across sequence lengths on AMA. Neu-PiG maintains accuracy and stability as duration increases.}
    \label{fig:sequence_length}
\end{figure*}

\paragraph{Evaluation Metrics.}
We assess reconstruction quality using $ \ell_2 $-Chamfer Distance (CD), Normal Consistency (NC), F-score, and Correspondence Error (Corr.). 
CD measures bidirectional surface deviations, while NC evaluates local smoothness and normal alignment. 
The Correspondence Error measures temporal consistency of vertex trajectories across frames and the F-score is chosen with a threshold of $0.5\%$.
We additionally report the average per-sequence runtime to evaluate computational efficiency.

\subsection{Quantitative Comparison}
We evaluate Neu-PiG on the DFAUST~\cite{dfaust}, AMA~\cite{ama}, and DT4D~\cite{dt4d} benchmarks, comparing it against both learning-based and training-free baselines (see \cref{tab:metrics}). %
Overall, both variants achieve strong performance across all datasets.
Although the longer schedule increases computation time, the resulting accuracy gains are minor. On DFAUST, the differences are marginal, while the others benefit only from slight refinement.
This behavior aligns with the convergence trends shown in \cref{fig:reconstruction_length}, where Neu-PiG yields high-quality reconstructions within a few seconds.

Across all benchmarks, Neu-PiG achieves the lowest Chamfer Distance and Correspondence Error, as well as the highest Normal Consistency and F-score, while being more than 60$\times$ faster than prior training-free approaches.
It consistently outperforms all training-free baselines on human motion datasets and maintains high fidelity on the challenging DT4D animal sequences (see \cref{fig:dt4d}), demonstrating reliable generalization across diverse motion types.


%
%

\paragraph{Scalability.}
Maintaining consistent correspondences over long and complex sequences is a persistent  challenge in dynamic surface reconstruction, as errors tend to accumulate with sequence length. 
To evaluate scalability, we split the AMA dataset into sequences ranging from 40 to 120 frames and report the performance in \cref{tab:length,fig:sequence_length}. 
While PDG requires progressively longer runtimes and ultimately fails on very long sequences and DynoSurf shows pronounced degradation, Neu-PiG reconstructs complete sequences in under two minutes while maintaining stable correspondences and high geometric fidelity.

\subsection{Ablation Study}
We perform ablations to assess the impact of key design choices, including architectural components, loss weighting and temporal frequency encoding.
Additional ablation results are provided in the supplementary material.

\paragraph{Method Components.}
We evaluate the contribution of individual components by selectively disabling or modifying parts of Neu-PiG.
Specifically, we remove the normal-direction latent encoding, omit preconditioning, replace the multi-resolution grid with the hash encoding of \cite{muller2022instant}, and test a single-resolution variant matching the finest grid parameters.
Each modification degrades reconstruction quality, as shown in \cref{tab:method}, confirming that the full combination of components yields the best overall performance.
In particular, hash grids are more memory-efficient but sacrifice spatial smoothness, which is important in our setting for coherent latent codes and stable deformations.

\begin{table}
    \scriptsize
    \centering
    \caption{
    Scalability on AMA sequences of increasing length. \mbox{Neu-PiG} remains stable and efficient, unlike PDG and DynoSurf.}
    \renewcommand{\arraystretch}{1.1}
    \setlength{\tabcolsep}{4.5pt}
    \begin{tabular}{c|l|ccccc}
        \toprule
        $ T $ & & CD $ [\times 10^{-5} ]$ $\downarrow$ & NC $\uparrow$  & F-$0.5\%$ $\uparrow$  & Corr. $\downarrow$ &Time $\downarrow$\\ 
        \midrule
        \multirow{3}{*}{{40}}
        & DynoSurf & \phantom{0}4.64 & 0.857 & 0.686 & 0.096 & \phantom{0}35\,\text{min} \\
        & PDG & \phantom{0}0.66 & 0.923 & 0.971 & 0.042 & \phantom{0}28\,\text{min} \\
        & Ours & \textbf{\phantom{0}0.53} & \textbf{0.947} & \textbf{0.981} & \textbf{0.019} & \textbf{\phantom{0}47\,\text{s}\phantom{mi}} \\
        \midrule
        \multirow{3}{*}{{60}}
        & DynoSurf & 14.64 & 0.778 & 0.474 & 0.132 & \phantom{0}48\,\text{min} \\
        & PDG & \phantom{0}0.72 & 0.919 & 0.960 & 0.054 & \phantom{0}63\,\text{min} \\
        & Ours & \textbf{\phantom{0}0.60} & \textbf{0.945} & \textbf{0.976} & \textbf{0.020} & \textbf{\phantom{0}63\,\text{s}\phantom{mi}} \\
        \midrule
        \multirow{3}{*}{{80}}
        & DynoSurf & 16.32 & 0.759 & 0.431 & 0.163 & \phantom{0}60\,\text{min}  \\
        & PDG & \phantom{0}1.35 & 0.906 & 0.931 & 0.089 & \phantom{0}93\,\text{min} \\
        & Ours & \textbf{\phantom{0}1.04} & \textbf{0.940} & \textbf{0.959} & \textbf{0.023} & \textbf{\phantom{0}84\,\text{s}\phantom{mi}} \\
        \midrule
        \multirow{3}{*}{{100}}
        & DynoSurf & 26.47 & 0.706 & 0.354 & 0.133 & \phantom{0}74\,\text{min} \\
        & PDG & \phantom{0}8.59 & 0.853 & 0.720 & 0.091 & 124\,\text{min}  \\
        & Ours & \textbf{\phantom{0}0.95} & \textbf{0.936} & \textbf{0.933} & \textbf{0.024} & \textbf{\phantom{0}97\,\text{s}\phantom{mi}} \\
        \midrule
        \multirow{3}{*}{{120}}
        & DynoSurf & 39.00 & 0.673 & 0.284 & 0.140 & \phantom{0}96\,\text{min}  \\
        & PDG & 30.20 & 0.788 & 0.571 & 0.118 & 158\,\text{min}  \\
        & Ours & \textbf{\phantom{0}1.31} & \textbf{0.926} & \textbf{0.887} & \textbf{0.028} & \textbf{110\,\text{s}\phantom{mi}} \\
        \bottomrule
    \end{tabular}
    \label{tab:length}
\end{table}

\paragraph{Time Frequency Function.}
We investigate how different time-encoding formulations affect our temporal deformation model. 
This module maps the normalized timestep $ { \tilde{t} \in [0,1] } $ into a high-dimensional representation for the deformation network. 
We compare several time–frequency strategies, each transforming $\tilde{t}$ by a distinct function:
The first variant employs a \emph{polynomial basis}, a simple deterministic encoding that represents time using monomials,
\begin{equation}
    \vec{\gamma}_{\mathrm{polynomial}}(t) = \left[ \tilde{t}^j \right]_{j=1}^{2M}.
\end{equation}
The second uses a \emph{Gaussian Fourier mapping} with a standard normal distribution, $ { \vec{B} \sim \mathcal{N}(\vec{0},\mathbf{I}) } $,
\begin{equation}
    \vec{\gamma}_{\mathrm{gaussian}}(t) = \left[ \sin(2\pi \vec{B} \tilde{t}), \cos(2\pi \vec{B} \tilde{t}) \right]_{j=1}^{M}.
\end{equation}
Finally we consider a \emph{learned embedding}, obtained from a two-layer MLP $\psi_t$,
\begin{equation}
    \vec{\gamma}_{\mathrm{learned}}(t) = \psi_t(\tilde{t}) \in \mathbb{R}^{2M}.
\end{equation}
%
%
In addition, we compare these baselines against our proposed time-Fourier encoding (\cref{sec:temporal_deformation_model}). 
All variants are trained under identical settings, with results in \cref{tab:time_frequencies} showing that the Fourier formulation yields the best balance of reconstruction accuracy and temporal coherence.

\paragraph{Stability Function.}
\label{par:stability_function}
In \cref{sec:optimization_objectives}, we adopt the deformation stability scaling proposed in PDG~\cite{kaltheuner2025preconditioned}, which stabilizes optimization by gradually increasing the confidence in later timesteps.
While effective, this mechanism can also slow convergence when early-frame reconstructions remain inaccurate.
To assess its influence, we ablate both components of the stability term, the catch-up variable~$\delta$ and the temporal weighting~$\omega$ on sequences of 40 frames, with the results presented in \cref{tab:stability}.
For the catch-up variable~$\delta$, which controls the confidence increase over training epochs~$\tilde{e}$, we evaluate four schedules: \textit{Constant} ($ {\delta = 1 } $), \textit{linear} ($ {\delta = 1 - \tilde{e} } $), \textit{exponential} ($ {\delta = \mathrm{e}^{- c \, \tilde{e}} } $ with $ {c = 5 } $), \textit{interpolated} ($ {w_{\mathrm{conf}}(t) = (1 - \tilde{e}) \, \omega(t) + \tilde{e} } $).
%
In addition, we test simplified formulations of the temporal weighting~$\omega$ within the confidence function~$w_{\mathrm{conf}}(t)$.
First, we replace the cumulative temporal product with a direct weighting:
\begin{equation}
    w_{\mathrm{conf}}(t) = \omega(t)^\delta
\end{equation}
Then, we compare three alternative definitions of~$\omega$: \textit{Constant} ($ {\omega(t) = 1 } $), \textit{delta-based} ($ {\omega(t) = 0.5 } $), \textit{single} ($ {\omega(t) =  1 / (1 + \max(0, \mathrm{cd}_t - \mathrm{cd}_{t_{\mathrm{key}}})) } $).
It isolates the effect of each stabilization term, allowing us to quantify its impact on convergence speed and reconstruction accuracy.

\begin{table}
    \scriptsize
    \centering
    \caption{
    Ablation on architectural components. Removing any module reduces accuracy, confirming their complementary roles.}
    \renewcommand{\arraystretch}{1.1}
    \setlength{\tabcolsep}{4.5pt}
    \begin{tabular}{l|cccc}
        \toprule
          & CD $ [\times 10^{-5} ] $ $\downarrow$ & NC $\uparrow$  & F-$0.5\%$ $\uparrow$  & Corr. $\downarrow$ \\
        \midrule
        Hash Encoding & 1.23 & 0.903 & 0.947 & 0.045 \\
        w/o Preconditioning & 0.98 & 0.955 & 0.958 & 0.036 \\
        w/o $\vec{z}_{\mathrm{n}}$ & 0.91 & 0.968 & 0.965 & 0.036 \\
        w/o $\mathcal{L}_{\mathrm{iso}}$ & 1.03 & 0.961 & 0.959 & 0.044 \\
        Single Level $ L = 1 $ & 0.98 & 0.965 & 0.961 & 0.036 \\
        Ours &  \textbf{0.87} & \textbf{0.969} & \textbf{0.968} & \textbf{0.034} \\
        \bottomrule
    \end{tabular}
    \label{tab:method}
\end{table}

\begin{table}
    \scriptsize
    \centering
    \caption{
    Comparison of time encoding strategies. Our Fourier encoding achieves the best accuracy and coherence.}
    \renewcommand{\arraystretch}{1.25}
    \begin{tabular}{l|cccc}
        \toprule
         & CD $ [\times 10^{-5} ]$ $\downarrow$ & NC $\uparrow$  & F-$0.5\%$ $\uparrow$  & Corr. $\downarrow$ \\
        \midrule
         Polynomial & 0.46 & \textbf{0.951} & 0.987 & 0.019 \\
         Gaussian & 0.77 & 0.945 & 0.979 & 0.046  \\
         Learned & 0.50 & 0.950 & 0.983 & 0.022 \\
         Ours & \textbf{0.44} & \textbf{0.951} & \textbf{0.989} & \textbf{0.017} \\
         \bottomrule
    \end{tabular}
    \label{tab:time_frequencies}
\end{table}

\begin{table}
    \scriptsize
    \centering
    \caption{
    Ablation of stability terms $\delta$ and $\omega$ on AMA with $ { T = 40 } $. Our formulation yields best convergence and accuracy.}
    \renewcommand{\arraystretch}{1.25}
    \begin{tabular}{l|l|cccc}
        \toprule
         &  & CD $ [\times 10^{-5} ]$ $\downarrow$ & NC $\uparrow$  & F-$0.5\%$ $\uparrow$  & Corr. $\downarrow$ \\
        \midrule
        \multirow{5}{*}{{$\delta$}}
         & Constant & 3.22 & 0.905 & 0.899 & 0.114 \\
         & Linear & 0.60 & 0.946 & 0.971 & 0.025  \\
         & Exponential & 0.53 & 0.946 & \textbf{0.982} & 0.030 \\
         & Interpolated & 0.64 & 0.945 & 0.972 & 0.046 \\
         & Ours & \textbf{0.53} & \textbf{0.947} & 0.981 & \textbf{0.025} \\
         \midrule
        \multirow{4}{*}{{$\omega(t)$}}
         & Constant & 3.27 & 0.905 & 0.898 & 0.114  \\
         & Delta-based & 2.96 & 0.909 & 0.908 & 0.112 \\
         & Single & 1.15 & 0.931 & 0.952 & 0.098 \\
         & Ours & \textbf{0.53} & \textbf{0.947} & \textbf{0.981} & \textbf{0.025} \\
        \bottomrule
    \end{tabular}
    \label{tab:stability}
\end{table}

\subsection{Limitations}
While our approach achieves fast and stable reconstructions, several practical limitations remain.
The method assumes a fixed topology derived from the keyframe mesh, which prevents recovery from incorrect or incomplete initial surfaces.
Moreover, the representational capacity of the latent grids and network bounds the sequence length that can be modeled effectively.
Finally, as correspondence is inferred implicitly through the Chamfer distance, very large motions or occlusions may reduce reconstruction fidelity, and in rare cases of extreme non-uniform deformation, such as strong local surface shrinkage, may also lead to local face flipping away from the keyframe.

%% file: sec/5_conclusion.tex
\section{Conclusion}

We presented Neu-PiG, a fast method for temporally consistent surface reconstruction from unstructured dynamic point clouds.
By encoding deformations across all time steps into a preconditioned multi-scale latent grid parameterized by a canonical surface, our method achieved spatially smooth and temporally stable reconstructions without explicit correspondences or learned priors.
Experiments demonstrated that our method attains higher fidelity and stability than existing state-of-the-art approaches while running over an order of magnitude faster.
We believe Neu-PiG offers a scalable foundation for future research in real-time 4D reconstruction and dynamic scene understanding.

\subsection*{Acknowledgements}
This research has been funded by the Federal Ministry of Educa-
tion and Research under grant no. 01IS22094A WEST-AI, by the
Federal Ministry of Education and Research of Germany as well as
the state of North-Rhine Westphalia as part of the Lamarr-Institute
for Machine Learning and Artificial Intelligence, by the Ministry of
Culture and Science North Rhine-Westphalia under grant number
PB22-063A (InVirtuo 4.0: Experimental Research in Virtual Envi-
ronments) and by the state of North Rhine-Westphalia as part of
the Excellency Start-up Center.NRW (U-BO-GROW) under grant
number 03ESCNW18B. 

%% file: sec/X_suppl.tex
\clearpage
\setcounter{page}{1}
\maketitlesupplementary
In this supplementary document, we provide additional experiments, analyses, and visualizations that complement the results presented in the main paper. 
Specifically, we extend the evaluation of Neu-PiG with further quantitative and qualitative results to validate the impact of grid design choices, preconditioning strength, and input conditions.

\begin{figure*}
    \centering
    \includegraphics[width=\linewidth, trim={0cm 0cm 0cm 0cm},clip]{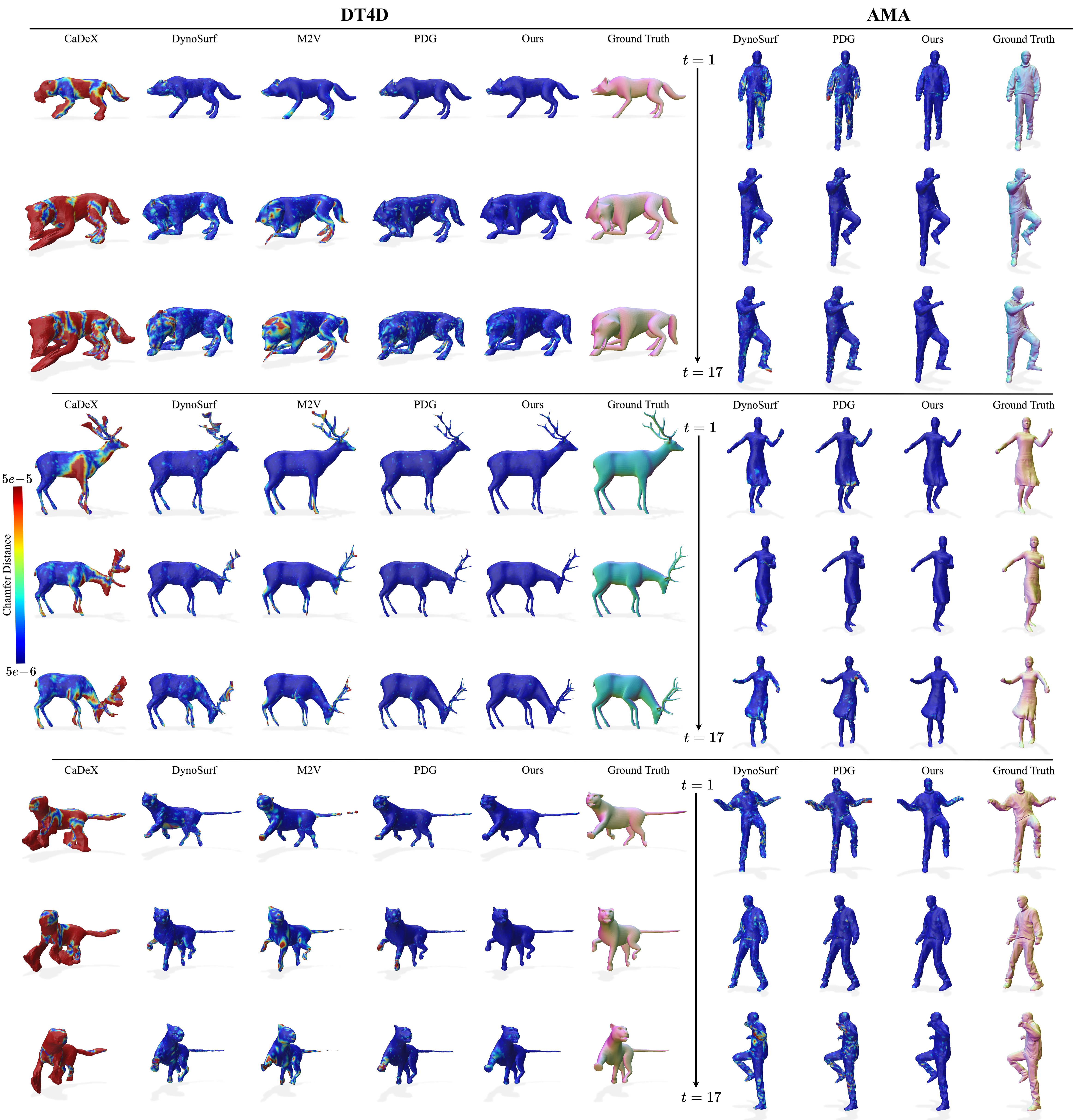}
    \caption{Extended qualitative results across diverse human and animal sequences. 
    Neu-PiG consistently reconstructs temporally stable and detailed surfaces, even under large non-rigid deformations and challenging motion dynamics. 
    }
    \label{fig:more_results}
\end{figure*}

\begin{figure*}
    \centering
    \includegraphics[width=\linewidth, trim={0cm 0cm 0cm 0cm},clip]{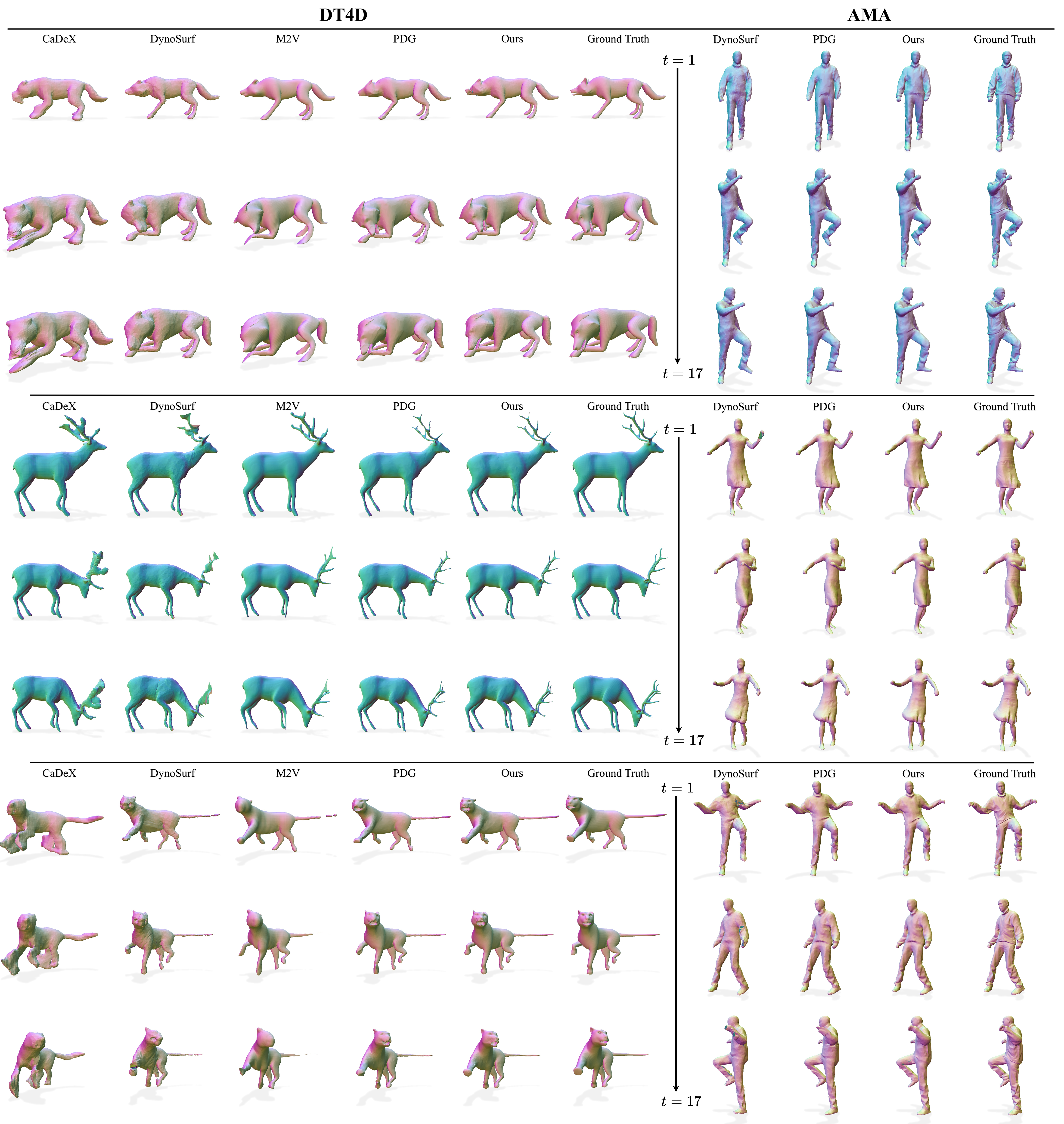}
    \caption{Extended visual results across diverse human and animal sequences. 
    The surface normals of the deformed meshes are shown in comparison to ground truth surface normals.
    }
    \label{fig:more_results_normal}
\end{figure*}

In \cref{fig:more_results,fig:more_results_normal}, we present extended visual results on a broader range of datasets, demonstrating that Neu-PiG maintains high reconstruction quality and temporal consistency across diverse motion types, outperforming existing state-of-the-art approaches in both accuracy and stability.

\subsection{Latent Grid Design}
We analyze the design choices of our preconditioned multi-resolution grid, focusing on the impact of the number of levels and the smoothness introduced by preconditioning.

\begin{figure}
    \centering
    \includegraphics[width=\linewidth, trim={1cm 3cm 0.5cm 4cm},clip]{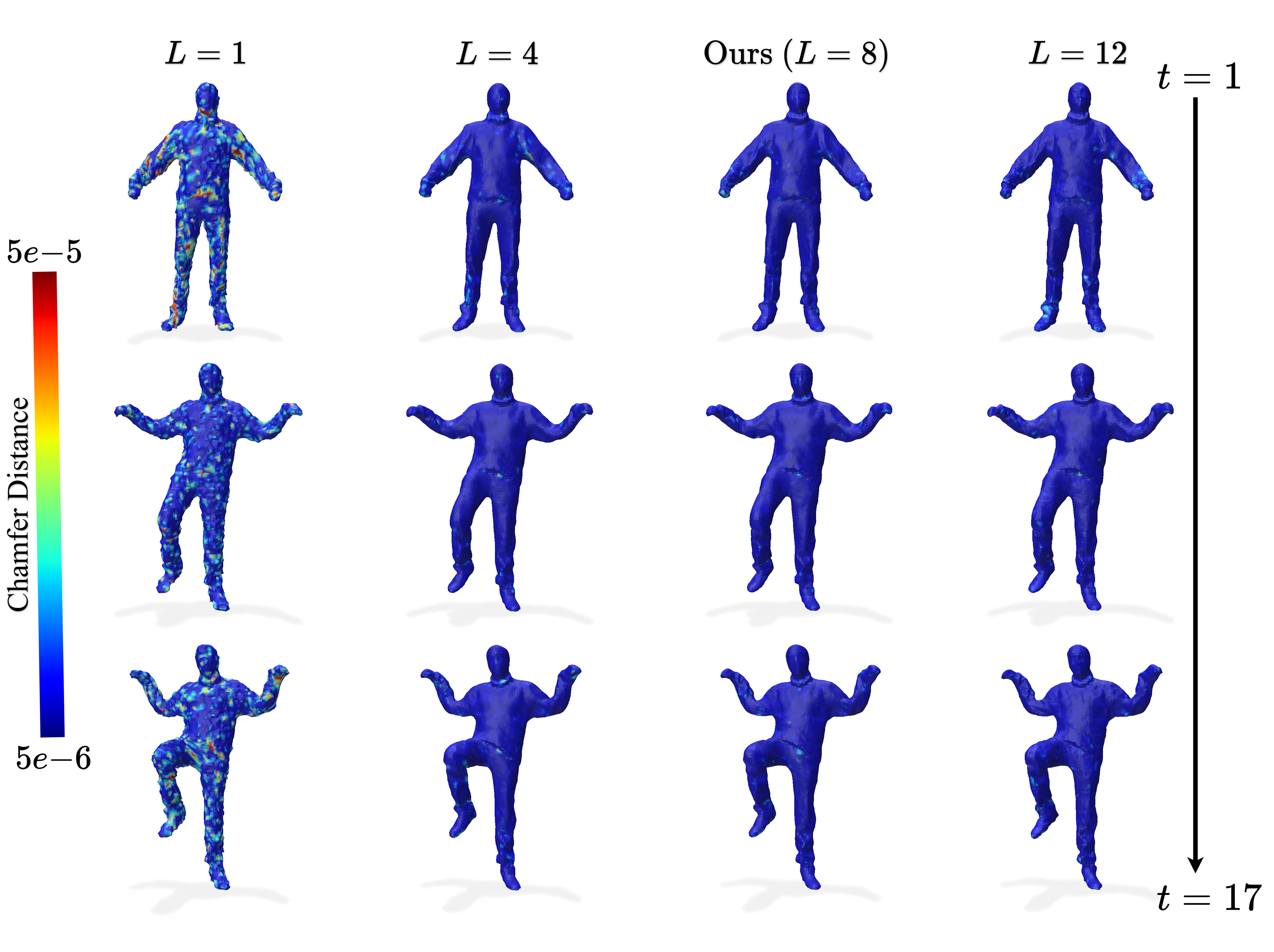}
    \caption{Qualitative analysis of the effect of the number of grid levels during optimization on reconstruction quality.
    Neu-PiG achieves accurate and temporally stable reconstructions even with few grid levels, while excessively increasing resolution leads to mild overfitting and loss of fine structural detail.}
    \label{fig:grid_level}
\end{figure}

\paragraph{Grid Levels.}
In \cref{tab:grid_lvl}, we evaluate the influence of the number of grid levels, described in \cref{sec:latent_grid}, on reconstruction quality using the AMA dataset.
We keep our default configuration and vary only the total number of levels $L$.
The results show that even with a small number of levels, and thus reduced latent capacity, Neu-PiG achieves surprisingly strong reconstructions.
However, excessively increasing the grid resolution leads to performance degradation due to overfitting, as shown in \cref{fig:grid_level}.

\begin{figure}
    \centering
    \includegraphics[width=\linewidth, trim={3cm 3cm 5.5cm 3cm},clip]{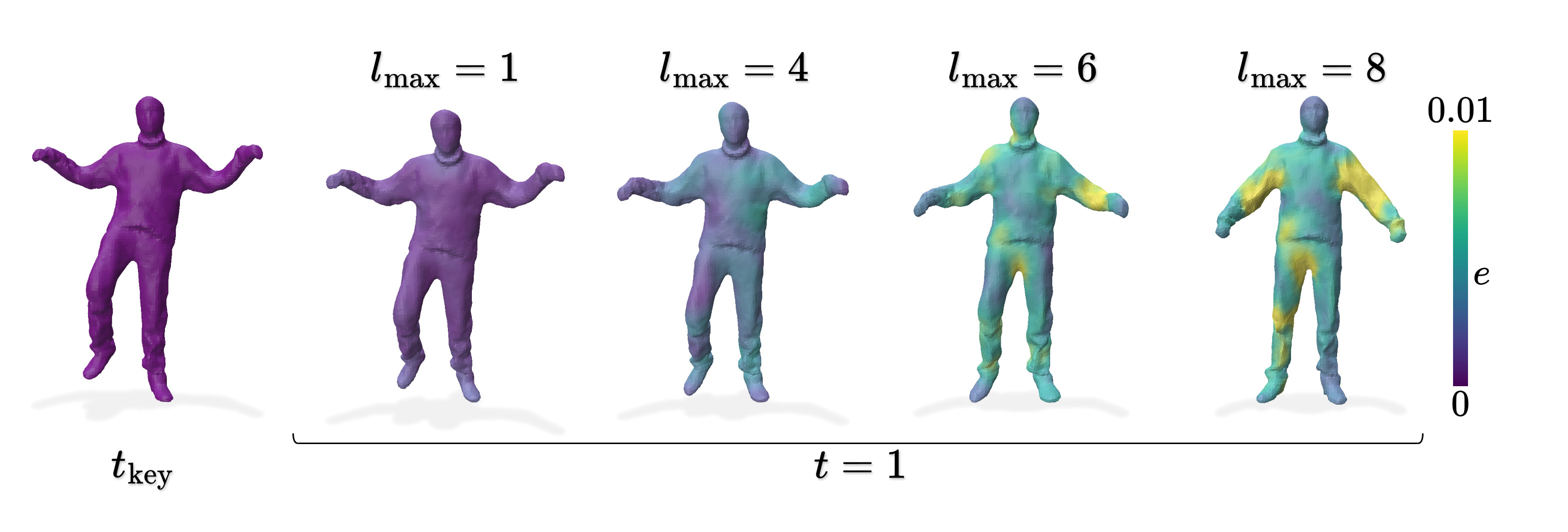}
    \caption{After optimization, we progressively restrict Neu-PiG to $l_\text{max}$ grid levels and visualize the deformations, colored by a rigidity proxy. It shows that coarse levels capture smooth, global motion, whereas finer levels contribute localized, high-frequency detail, supporting the role of the multi-resolution latent grid.}
    \label{fig:grid_level2}
\end{figure}

Complementing this quantitative analysis, \cref{fig:grid_level2} provides a qualitative view of the learned multi-resolution decomposition.
Coarser levels capture low-frequency global motion, whereas finer levels recover localized high-frequency deformations, visualized using a local rigidity proxy based on Kabsch alignment.

\begin{figure}
    \centering
    \includegraphics[width=\linewidth, trim={1cm 6cm 8cm 1cm},clip]{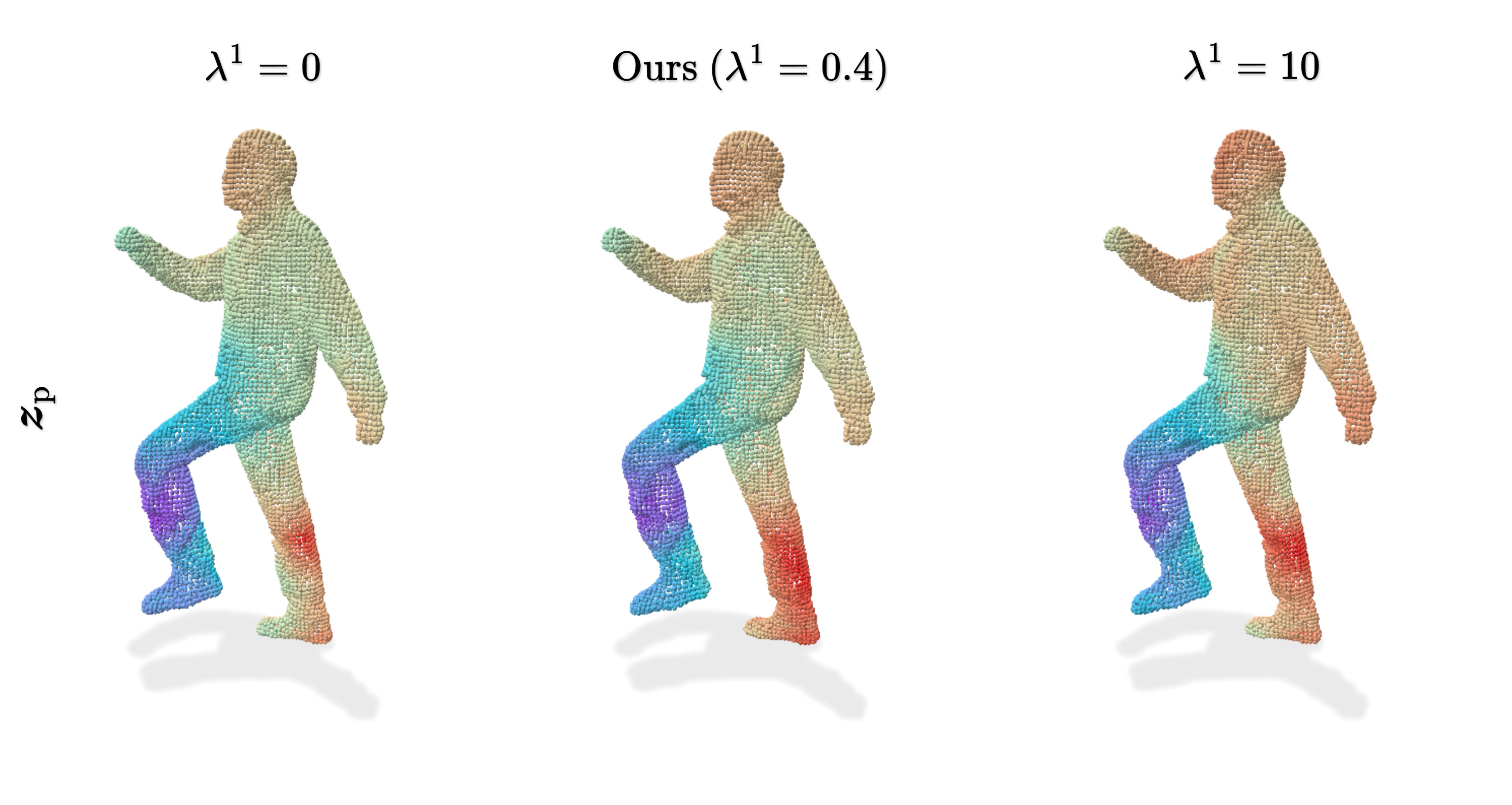}
    \caption{Visualization of the learned latent space for different smoothness weights.  
We perform a PCA analysis on the latent vectors and display the first principal component as a spatial field.}
    \label{fig:lambda_weight}
\end{figure}

\paragraph{Smoothness Weights.}
We investigate the effect of the smoothness parameter $\lambda$ in our grid preconditioning, introduced in \cref{sec:optimization_objectives}, on reconstruction performance.
To access its influence, we vary the base smoothness value $\lambda^1$ to both smaller and larger magnitudes, while using the same increase per level as in the default configuration.
The results in \cref{tab:lambda_weight} on the AMA dataset indicate that Neu-PiG is largely robust to moderate changes in smoothness, while extreme values, either excessively high or entirely absent, lead to a clear degradation in reconstruction quality.
We further visualize the learned latent spaces in \cref{fig:lambda_weight} by performing a PCA analysis and plotting the first principal component.

\begin{table}
    \scriptsize
    \centering
    \caption{
    Ablation study on the number of grid levels in $\set{G}_{\mathrm{p}}$, evaluated on the AMA dataset.
    Increasing the number of levels improves reconstruction quality up to ${ L=8 } $, after which performance slightly degrades due to overfitting.}
    \renewcommand{\arraystretch}{1.25}
    \begin{tabular}{l|cccc}
        \toprule
         & CD $ [\times 10^{-5} ]$ $\downarrow$ & NC $\uparrow$  & F-$0.5\%$ $\uparrow$  & Corr. $\downarrow$ \\
        \midrule
        $ L = 1 $ & 1.96 & 0.850 & 0.810 & 0.063 \\
        $ L = 2 $ & 1.76 & 0.859 & 0.841 & 0.054  \\
        $ L = 4 $ & 0.54 & 0.945 & 0.980 & 0.023 \\
        $ L = 6 $ & 0.46 & 0.949 & 0.987 & 0.018 \\
        Ours ($ L = 8 $) & \textbf{0.44} & \textbf{0.951} & \textbf{0.989} & \textbf{0.017} \\
        $ L = 10 $ & 0.45 & \textbf{0.951} & \textbf{0.989} & 0.021 \\
        $ L = 12 $ & 0.46 & 0.948 & 0.988 & 0.024 \\
        \bottomrule
    \end{tabular}
    \label{tab:grid_lvl}
\end{table}


\subsection{Input Conditions}
We additionally study how different input conditions, such as varying number of input points and noise levels, affect the reconstruction performance of Neu-PiG.

\paragraph{Noise.}
We evaluate the robustness of Neu-PiG to noisy input data by adding Gaussian noise with varying strengths to the input points based on the size of the bounding box diagonal.
As shown in \cref{tab:noiset}, the method remains stable for noise levels up to $0.25–0.5 \%$, after which the performance degrades as noise begins to be reconstructed as part of the surface geometry.

\paragraph{Point Cloud Resolution.}
We evaluate the reconstruction performance across varying input point cloud resolutions, focusing on the generalization behavior in comparison to DynoSurf, which similarly employs MLP-based transformation modeling. 
As shown in \cref{tab:metrics_res}, DynoSurf exhibits a noticeable drop in performance with increasing target resolution, indicating limited scalability. 
In contrast, our method demonstrates consistent improvements with higher input densities, closely matching the behavior of the direct optimization approach PDG, and achieving superior reconstruction quality at all tested resolutions.
The difference compared to DynoSurf is particularly noteworthy: although both methods employ MLPs to model per-point transformations, DynoSurf fails to benefit from denser point clouds.
This robustness stems from our hierarchical latent representation, which efficiently captures both coarse and fine geometric structures.

\begin{table}
    \scriptsize
    \centering
    \caption{
    Ablation study on the smoothness weight $\lambda$ of the base grid level ($ { l=1 } $), evaluated on the AMA dataset.
    Neu-PiG remains robust under moderate variations of $\lambda$, while extreme values reduce reconstruction quality.}
    \renewcommand{\arraystretch}{1.25}
    \begin{tabular}{l|cccc}
        \toprule
         & CD $ [\times 10^{-5} ]$ $\downarrow$ & NC $\uparrow$  & F-$0.5\%$ $\uparrow$  & Corr. $\downarrow$ \\
        \midrule
        $ \lambda^1 = 0 $ & 0.58 & 0.928 & 0.977 & 0.022 \\
        $ \lambda^1 = 0.08 $ & 0.45 & 0.950 & 0.988 & 0.020 \\
        Ours ($ \lambda^1 = 0.4 $) & \textbf{0.44} & \textbf{0.951} & \textbf{0.989} & \textbf{0.017} \\
        $ \lambda^1 = 2 $ & 0.46 & 0.948 & 0.986 & 0.021 \\
        $ \lambda^1 = 10 $ & 0.50 & 0.946 & 0.983 & 0.025 \\
        \bottomrule
    \end{tabular}
    \label{tab:lambda_weight}
\end{table}

\subsection{Neural Model}
We further analyze the design of the neural components in Neu-PiG and evaluate how different architectural choices of the decoder influence reconstruction quality and temporal stability.

\paragraph{Time Deformation Model.}
We investigate how modifications to the temporal deformation network influence reconstruction performance.
This component models temporal shape evolution conditioned on the latent embeddings introduced in \cref{sec:latent_grid}.
We systematically vary the dimensionality of the positional and normal latent vectors $\vec{z}_{\mathrm{p}}$ and $\vec{z}_{\mathrm{n}}$, as well as the frequency range of the time encoding $ \vec{\gamma} $.
In addition, we assess the impact of network capacity by adjusting the number of hidden units per layer.
As shown in \cref{tab:deformation_model}, Neu-PiG remains stable across most configurations.
However, excessive reduction in model capacity leads to lower reconstruction quality, while increasing the dimensionality of $\vec{z}_{\mathrm{p}}$ or $\vec{\gamma}$ causes the network to overfit more easily.

\paragraph{Rotation Modeling.}
Our deformation model predicts parameters that are mapped to 3D rotations and deformations, as described in \cref{sec:transformation_mapping}.
Several representations exist for mapping parameters to rotations, including quaternions, the Cayley transform, and the exponential map, and their choice can affect optimization stability, convergence behavior, and final reconstruction accuracy.
To study this influence, we run our reconstruction using each representation under identical settings and compare performance over different amounts of maximal optimization epochs.
The resulting metrics are summarized in \cref{tab:rotation}.
The results show that while all parameterizations achieve comparable accuracy, quaternions converge slightly faster and yield marginally better performance during early training.

\begin{table}
    \scriptsize
    \centering
    \caption{
    Ablation study on robustness to input noise with varying strengths based on the bounding box diagonal.
    Neu-PiG maintains stable reconstruction quality up to $0.5\%$ noise, after which performance degrades as noise is captured in the surface geometry.}
    \renewcommand{\arraystretch}{1.25}
    \begin{tabular}{l|cccc}
        \toprule
         Noise & CD $ [\times 10^{-5} ]$ $\downarrow$ & NC $\uparrow$  & F-$0.5\%$ $\uparrow$  & Corr. $\downarrow$ \\
        \midrule
        $0.25\%$ & 0.69 & 0.925 & 0.967 & 0.021 \\
        $0.5\%$ & 1.34 & 0.866 & 0.853 & 0.026 \\
        $1\%$ & 4.45 & 0.681 & 0.540 & 0.049 \\
        $2\%$ & 21.71 & 0.517 & 0.247 & 0.086 \\
        \bottomrule
    \end{tabular}
    \label{tab:noiset}
\end{table}

\begin{table}
    \scriptsize
    \centering
    \caption{
    Reconstruction performance at varying input point cloud resolutions.
    Neu-PiG consistently improves with higher input densities, closely matching the behavior of PDG while maintaining superior quality across all settings.
    In contrast, DynoSurf shows limited scalability and degrading performance at higher resolutions.}
    \renewcommand{\arraystretch}{1.25}
    \begin{tabular}{c|l|cccc}
        \toprule
        $ \abs{\set{P}_t} $ & & CD $ [\times 10^{-5} ] $ $\downarrow$ & NC $\uparrow$  & F-$0.5\%$ $\uparrow$  & Corr. $\downarrow$ \\
        \midrule
        \multirow{3}{*}{{\phantom{0}2500}}
        & DynoSurf & 1.56 & 0.896 & 0.862 & 0.041 \\
        & PDG & 0.79 & 0.922 & 0.948 & 0.037  \\
        & Ours & \textbf{0.75} & \textbf{0.930} & \textbf{0.953} & \textbf{0.022}  \\
        \midrule
        \multirow{3}{*}{{\phantom{0}5000}}
        & DynoSurf & 1.01 & 0.918 & 0.921 & 0.044 \\
        & PDG & 0.47 & 0.939 & 0.985 & 0.030 \\
        & Ours & \textbf{0.44} & \textbf{0.950} & \textbf{0.988} & \textbf{0.018}  \\
        \midrule
        \multirow{3}{*}{{10000}}
        & DynoSurf & 1.28 & 0.906 & 0.897 & 0.037 \\
        & PDG & 0.40 & 0.950 & 0.993 & 0.027 \\
        & Ours & \textbf{0.37} & \textbf{0.960} & \textbf{0.995} & \textbf{0.015}  \\
        \midrule
        \multirow{3}{*}{{20000}}
        & DynoSurf & 1.45 & 0.902 & 0.887 & 0.040 \\
        & PDG & 0.37 & 0.959 & 0.996 & 0.023 \\
        & Ours & \textbf{0.34} & \textbf{0.968} & \textbf{0.997} & \textbf{0.014}  \\
        \bottomrule
    \end{tabular}
    \label{tab:metrics_res}
\end{table}

\begin{table}
    \scriptsize
    \centering
    \caption{
    Ablation study on the temporal deformation network architecture, evaluated on the AMA dataset with 40 time steps per sequence.
    We vary latent dimensionality, number of time-encoding frequencies, and network capacity to assess their effect on reconstruction quality.
    Neu-PiG remains stable under most configurations, though extreme reductions in capacity or excessive latent dimensionality lead to degraded performance or overfitting.
    Our default configuration is shown in the last row.}
    \renewcommand{\arraystretch}{1.25}
    \setlength{\tabcolsep}{4.5pt}
    \begin{tabular}{cccc|cccc}
        \toprule
        $\abs{\psi}$ & $\abs{\vec{z}_{\mathrm{p}}}$ & $\abs{\vec{z}_{\mathrm{n}}}$ & $\abs{\vec{\gamma}}$ & CD $ [\times 10^{-5} ]$ $\downarrow$ & NC $\uparrow$  & F-$0.5\%$ $\uparrow$   & Corr. $\downarrow$\\
        \midrule
        512 & 30 & 2 & 2 & 0.63 & 0.945 & 0.970 & 0.027 \\
        512 & 30 & 2 & 32 & 0.64 & 0.943 & 0.973 & 0.040 \\
        512 & 30 & 1 & 8 & 0.55 & 0.946 & 0.979 & 0.026 \\
        512 & 30 & 8 & 8 & 0.57 & \textbf{0.947} & 0.978  & 0.026 \\
        512 & 8 & 2 & 8 & 0.61 & 0.945 & 0.975 & 0.023 \\
        512 & 120 & 2 & 8 & 0.59 & \textbf{0.947} & 0.978 & 0.032 \\
        128 & 30 & 2 & 8 & 0.76 & 0.938 & 0.959 & 0.045 \\
        2048 & 30 & 2 & 8 & 0.54 & 0.948 & 0.980 & 0.021 \\
        512 & 30 & 2 & 8 & \textbf{0.53} & \textbf{0.947} & \textbf{0.981} & \textbf{0.019} \\
        \bottomrule
    \end{tabular}
    \label{tab:deformation_model}
\end{table}

\begin{table}
    \scriptsize
    \centering
    \caption{
    Comparison of different rotation parameterizations predicted by the deformation model.
    We evaluate quaternions, the Cayley transform, and the exponential map under identical optimization conditions.
    All representations achieve similar accuracy, while using quaternions leads to slightly faster convergence and improves performance with fewer epochs.}
    \renewcommand{\arraystretch}{1.25}
    \begin{tabular}{c|l|cccc}
        \toprule
        Epochs & & CD $ [\times 10^{-5} ]$ $\downarrow$ & NC $\uparrow$  & F-$0.5\%$ $\uparrow$  & Corr. $\downarrow$ \\
        \midrule
        \multirow{3}{*}{{75}}
        & Cayley & 1.96 & 0.921 & 0.866 & 0.047 \\
        & Exponential & 1.62 & \textbf{0.924} & \textbf{0.886} & \textbf{0.043}  \\
        & Quaternions & \textbf{1.59} & 0.923 & 0.884 & 0.044 \\
        \midrule
        \multirow{3}{*}{{125}}
        & Cayley & 0.72 & 0.940 & 0.955 & 0.029 \\
        & Exponential & 0.68 & 0.941 & 0.961 & \textbf{0.027}  \\
        & Quaternions & \textbf{0.65} & \textbf{0.942} & \textbf{0.965} & 0.028 \\
        \midrule
        \multirow{3}{*}{{250}}
        & Cayley & 0.52 & 0.947 & 0.978 & 0.021 \\
        & Exponential & 0.54 & \textbf{0.948} & 0.976 & 0.021  \\
        & Quaternions & \textbf{0.50} & \textbf{0.948} & \textbf{0.982} & \textbf{0.020} \\
        \midrule
        \multirow{3}{*}{{500}}
        & Cayley & 0.47 & \textbf{0.950} & 0.985 & 0.019 \\
        & Exponential & 0.47 & \textbf{0.950} & 0.985 & \textbf{0.018}  \\
        & Quaternions & \textbf{0.46} & \textbf{0.950} & \textbf{0.986} & \textbf{0.018} \\
        \midrule
        \multirow{3}{*}{{1000}}
        & Cayley & 0.46 & 0.950 & 0.987 & 0.019 \\
        & Exponential & 0.45 & \textbf{0.951} & \textbf{0.988} & \textbf{0.017}  \\
        & Quaternions & \textbf{0.44} & \textbf{0.951} & \textbf{0.988} &\textbf{0.017} \\
        \bottomrule
    \end{tabular}
    \label{tab:rotation}
\end{table}